\documentclass[10pt,twocolumn,letterpaper]{article}

\usepackage{cvpr}              

\usepackage[utf8]{inputenc} 
\usepackage[T1]{fontenc}    

\usepackage{graphicx}
\usepackage{amsmath}
\usepackage{amssymb}
\usepackage{booktabs}
\usepackage{graphicx}
\usepackage{subcaption}
\usepackage{dblfloatfix}
\usepackage{bm}
\usepackage{subcaption}
\usepackage{xcolor}

\usepackage[pagebackref,breaklinks,colorlinks]{hyperref}

\usepackage[capitalize]{cleveref}
\crefname{section}{Sec.}{Secs.}
\Crefname{section}{Section}{Sections}
\Crefname{table}{Table}{Tables}
\crefname{table}{Tab.}{Tabs.}

\def\cvprPaperID{36} 
\def\confName{CVPR}
\def\confYear{2023}

\begin{document}

\definecolor{darkgreen}{rgb}{0., 0.5, 0.}
\newcommand{\michal}[1]{\textcolor{darkgreen}{\small [mz: #1]}}
\newcommand{\kamil}[1]{\textcolor{orange}{\small [KD: #1]}}
\newcommand{\tomek}[1]{\textcolor{red}{\small [TT: #1]}}

\title{Exploring Continual Learning of Diffusion Models}

\author{Michał Zając\\
Jagiellonian University
\and
Kamil Deja\\
Warsaw University of Technology\\
\and
Anna Kuzina\\
Vrije Universiteit Amsterdam
\and
Jakub M. Tomczak\\
Eindhoven University of Technology
\and
Tomasz Trzciński\\
Warsaw University of Technology\\
Ideas NCBR\\
Tooploox
\and
Florian Shkurti\\
University of Toronto\\
Vector Institute
\and
Piotr Miłoś\\
Ideas NCBR\\
Polish Academy of Sciences
}

\maketitle

\begin{abstract}

Diffusion models have achieved remarkable success in generating high-quality images thanks to their novel training procedures applied to unprecedented amounts of data. However, training a diffusion model from scratch is computationally expensive. This highlights the need to investigate the possibility of training these models iteratively, reusing computation while the data distribution changes. In this study, we take the first step in this direction and evaluate the continual learning (CL) properties of diffusion models.
We begin by benchmarking the most common CL methods applied to Denoising Diffusion Probabilistic Models (DDPMs), where we note the strong performance of the experience replay with the reduced rehearsal coefficient. Furthermore, we provide insights into the dynamics of forgetting, which exhibit diverse behavior across diffusion timesteps. We also uncover certain pitfalls of using the bits-per-dimension metric for evaluating CL.
\end{abstract}


\section{Introduction}

Diffusion models \cite{sohl2015deep} have recently gained popularity due to their state-of-the-art performance in generative image modeling. Models like DALL-E~2 \cite{dalle2}, and Imagen \cite{imagen} have produced unparalleled quality and diversity in generated images, inspiring the community to explore various creative applications. Stable training of Deep Diffusion Probabilistic Models (DDPMs) with unprecedented volumes of data is one of the key factors that enable these remarkable applications. For instance, the DALL-E 2 model was trained using approximately 650 million images \cite{dalle2}, and open-source implementations of this method have reported a training time of 56 days using TPU v3 hardware \cite{Dayma_DALLE_Mini_2021}. Such computational resources are beyond the reach of many academic institutions, smaller companies, and independent researchers. Therefore, reducing the computational burden of diffusion models is crucial to increase their accessibility and democratization. 
One promising approach for reducing resource requirements is to continually reuse previously trained models and train them with continuously incoming data.

Continual learning (CL) \cite{de2019continual,DBLP:journals/nn/ParisiKPKW19} is focused on the effectiveness of model training when data portions known as tasks are presented to the learner in a sequence, with a possibility of the changing training data distribution. There are several common goals for CL algorithms, such as preventing forgetting, increasing transfer, and limiting computational resources. Although most applications consider the problem of continual  supervised learning, there is a growing field of research focusing on CL in generative modeling \cite{lesort2019generative,zhai2019lifelong}. Nevertheless, to the best of our knowledge, diffusion models have yet to be tested in the continual learning setup.

We take the first step to fill this gap by investigating how diffusion models forget and how they behave when trained with different CL methods. To achieve this, we conduct a series of experiments on common MNIST \cite{deng2012mnist}, and Fashion-MNIST \cite{xiao2017fashion} datasets.
Firstly, we demonstrate that DDPMs suffer from catastrophic forgetting, which is evident in the reduced quality of generations from previous tasks when retrained with additional data. Next, we retrain the DDPM using the simple and effective Experience Replay \cite{chaudhry2019tiny} algorithm and confirm its ability to prevent catastrophic forgetting, working especially well with reduced relative weight for rehearsal loss.
In addition to our benchmarking experiments, we note and quantify some interesting phenomena. Firstly, we observe that the standard bits-per-dimension (BPD) metric does not capture forgetting well, as it can deteriorate only slightly despite the complete loss of generative ability. Secondly, we demonstrate the effect of timestep-dependent overfitting on the buffer data, consistent with the generation-denoising decomposition of DDPMs from \cite{deja2022analyzing}.



\section{Background and related work}

\subsection{Diffusion models}

In this section, we briefly introduce diffusion generative models \cite{sohl2015deep}, specifically Denoising Diffusion Probabilistic Models (DDPMs), as described in \cite{ho2020denoising,nichol2021improved}. The forward diffusion process gradually adds Gaussian noise to data samples $x_0$ from a data distribution $q(x_0)$, given by:
$$ q(x_t | x_{t-1}) = \mathcal{N}(x_t; \sqrt{1 - \beta_t}x_{t-1}, \beta_t \mathbb{I}). $$
The values of $\beta_t$ define a variance schedule that is set so that the resulting distribution becomes indistinguishable from an isotropic Gaussian after $\tau$ timesteps. The idea behind the diffusion models is to model the reverse diffusion process distributions $q(x_{t-1} | q_t)$ with a powerful approximator such as a deep neural network with parameters $\theta$:
$$ p_{\theta}(x_{t-1} | x_t) = \mathcal{N}(x_{t-1} ; \mu_{\theta}(x_t, t), \Sigma_{\theta}(x_t, t)). $$
This modeling approach allows us to start from the isotropic Gaussian and then go through the reverse diffusion process to model the data distribution. 

Training loss is constructed as a variational upper bound on the negative likelihood $\mathbb{E}[- \log p_{\theta} (x_0)]$, which can be written in the following form:

\begin{equation}
\label{eq:sum}
\begin{split}
\mathcal{L} := \underbrace{-\log p_\theta(x_0|x_1)}_{\mathcal{L}_0} + \underbrace{D_{KL}(q(x_T | x_0) || p(x_T))}_{\mathcal{L}_\tau} + \\ 
\sum_{t=2}^{\tau} \underbrace{D_{KL}(q(x_{t-1} | x_t, x_0) || p_{\theta}(x_{t-1} | x_t))}_{\mathcal{L}_{t-1}}.
\end{split}
\end{equation}
Individual parts of the loss can be computed analytically to yield
\begin{equation}
\label{eq:chunk}
\begin{split}
\mathcal{L}_t = \mathbb{E}_{x_0, \boldsymbol{\epsilon}} 
\bigg[ \frac{\beta_{t}^{2}}{2 \sigma_{t}^{2} \alpha_{t}\left(1-\overline{\alpha}_{t}\right)} \cdot \\ || \boldsymbol{\epsilon}-\boldsymbol{\epsilon}_{\theta}\left(\sqrt{\overline{\alpha}_{t}} x_0+\sqrt{1-\overline{\alpha}_{t}} \boldsymbol{\epsilon}, t\right) ||^{2} \bigg],
\end{split}
\end{equation}
where $\alpha_t := 1 - \beta_t, \overline{\alpha}_t := \prod_{s=0}^t \alpha_s, \boldsymbol{\epsilon} \sim \mathcal{N}(\boldsymbol{0}, \boldsymbol{1}), $ and $\boldsymbol{\epsilon}_{\theta}$ is the prediction of the noise that was used in the diffusion process.
A simplified form of the loss introduced by~\cite{ho2020denoising} omits the  multiplicative factor:
\begin{equation}
\label{eq:loss_simplified}
\mathcal{L}_t^{\text{simple}} = \mathbb{E}_{x_0, \boldsymbol{\epsilon}} 
\bigg[ || \boldsymbol{\epsilon}-\boldsymbol{\epsilon}_{\theta}\left(\sqrt{\overline{\alpha}_{t}} x_0+\sqrt{1-\overline{\alpha}_{t}} \boldsymbol{\epsilon}, t\right) ||^{2} \bigg].
\end{equation}

The model is trained using stochastic gradient descent, where the training data $x_0$, noise $\boldsymbol{\epsilon}$, and timestep $t$ are randomly selected, and either $\mathcal{L}_t$ or $\mathcal{L}_t^{\text{simple}}$ is used as the loss function.

\subsection{Continual learning}

Continual learning (CL) \cite{de2019continual,DBLP:journals/nn/ParisiKPKW19} pertains to the scenario where the data distribution changes during the learning process. CL methods aim to reduce forgetting of previous tasks and maximize forward transfer, which involves leveraging past tasks to facilitate faster and improved learning.

This paper examines continual learning for generative modeling, in which a sequence of $N$ tasks $T_1, T_2, \ldots, T_N$ is considered. Each task $T_i$ is associated with a dataset $\mathcal{D}_i$, and the objective is to model the distribution of $\mathcal{D}_i$, as measured by some metric $m$. We adopt standard CL metrics, namely average performance and forgetting. Specifically, we denote $m_{i,j}$ as the value of the metric $m$ for task $i$ after training on task $j$. The \emph{average $m$} is defined as the average value of $m$ across all tasks after training, i.e.,
$$\text{avg } m := \frac{1}{N}\sum_{i=1}^{N} m_{i,N}.$$
We also define $m$-\emph{forgetting} as
$$m\text{-forgetting} := \frac{1}{N}\sum_{i=1}^{N} (m_{i,N} - m_{i, i}).$$

\begin{table*}
    \caption{Numerical comparison of CL methods on MNIST dataset. We report means and standard errors over $10$ random seeds.}
    \label{tab:benchmark_mnist}
    \centering
\begin{tabular}{l|ll|ll}
\toprule
                             & avg FID 	$\downarrow$                                      & FID-forgetting  	$\downarrow$                         & avg BPD      	$\downarrow$                    & BPD-forgetting      	$\downarrow$          \\
\midrule
 \bf{Finetuning}         & $61.97 \pm 0.69$ & $55.59 \pm 0.76$ & $2.13 \pm 0.01$ & $0.10 \pm 0.01$  \\
 \bf{L2}                  & $46.02 \pm 0.72$ & $13.11 \pm 0.68$ & $1.95 \pm 0.00$ & $\bm{-0.01 \pm 0.01}$ \\
 \bf{Exp. replay, coef. = 1}  & $15.56 \pm 0.21$ & $10.62 \pm 0.19$ & $2.64 \pm 0.04$ & $0.72 \pm 0.04$  \\
 \bf{Exp. replay, coef. = 0.01} & $\bm{6.37 \pm 0.27}$  & $\bm{1.00 \pm 0.29}$  & $\bm{1.92 \pm 0.00}$ & $\bm{0.00 \pm 0.00}$ \\
 \bf{Gen. replay, coef. = 1}  & $10.36 \pm 0.53$ & $5.35 \pm 0.46$  & $1.99 \pm 0.01$ & $0.06 \pm 0.00$  \\
\bottomrule
\end{tabular}
\end{table*}

\begin{table*}
    \caption{Numerical comparison of CL methods on Fashion-MNIST dataset. We report means and standard errors over $10$ random seeds.}
    \label{tab:benchmark_fmnist}
    \centering
\begin{tabular}{l|ll|ll}
\toprule
                             & avg FID 	$\downarrow$                                      & FID-forgetting  	$\downarrow$                         & avg BPD      	$\downarrow$                    & BPD-forgetting      	$\downarrow$          \\
\midrule
 \bf{Finetuning}         & $107.53 \pm 0.70$ & $94.57 \pm 0.76$ & $3.62 \pm 0.01$ & $0.28 \pm 0.00$  \\
 \bf{L2}                 & $95.46 \pm 1.82$  & $\bm{-1.76 \pm 0.64}$ & $3.54 \pm 0.00$ & $\bm{-0.01 \pm 0.00}$ \\
 \bf{Exp. replay, coef. = 1}  & $31.34 \pm 0.53$  & $19.28 \pm 0.40$ & $4.04 \pm 0.01$ & $0.77 \pm 0.01$  \\
 \bf{Exp. replay, coef. = 0.01} & $\bm{18.27 \pm 0.86}$  & $5.36 \pm 0.99$  & $\bm{3.31 \pm 0.00}$ & $0.03 \pm 0.00$  \\
 \bf{Gen. replay, coef. = 1}  & $31.61 \pm 2.41$  & $19.60 \pm 1.89$ & $3.36 \pm 0.00$ & $0.08 \pm 0.00$  \\
\bottomrule
\end{tabular}
\end{table*}

As our metrics $m$, we use bits-per-dimension (BPD), which is appropriately normalized negative log-likelihood, and Frechet Inception Distance (FID) \cite{heusel2017gans}, a common metric capturing both quality and diversity of the samples from generative models. These metrics are standard in the literature on diffusion models \cite{ho2020denoising,nichol2021improved}.

In our experiments, we consider several standard CL methods. The first is \textbf{Finetuning}, where training is done sequentially on given tasks without particular adaptations.
Then, we consider two \emph{replay-based} methods that leverage data from past tasks in order to reduce forgetting. \textbf{Experience replay} \cite{chaudhry2019tiny} maintains a small buffer of examples for each past task and adds an auxiliary loss that is the same as the standard loss function but computed on the batch from replay buffers. \textbf{Generative replay} \cite{shin2017continual} is similar. However, instead of using buffers, it keeps a frozen version of the model from before the current task and uses it to generate rehearsal data from previous tasks. 
\emph{Regularization-based} methods constrain model parameters to stay close to the past ones, which worked well on previous tasks. In particular, the \textbf{L2} method adds the L2 distance between the current and historical network parameters as an auxiliary loss.

\subsubsection{Continual learning with generative models}

Generative models are typically used for CL in the generative rehearsal  \cite{2017shin+3, lesort2019generative} approach. This technique employs a generative model that generates rehearsal examples from previous tasks for the continually trained classifier. However, the generative model used for rehearsal may suffer from catastrophic forgetting, so it is often trained with a \emph{generative replay} that combines generations from previous tasks and new data samples.

Other approaches that focus on continual learning of generative models include extensions to regularization-based methods~\cite{nguyen2017variational}, buffer-based replay~\cite{rao2019continual}, or architectural adaptation with techniques such as hypernetworks~\cite{von2019continual}. Several works use the specific properties of Variational Autoencoders~\cite{kingma2014autoencoding} to continually align their latent space~\cite{ijcai2022p402} and shared features~\cite{achille2018life} or with additive aggregated posterior~\cite{egorov2021boovae}. Additionally, some works train GANs in the continual learning scenarios~\cite{wu2018memory}; \cite{ye2020learning} extends the approach to VAEGAN.

\section{Experiments}

In this section, we provide observations on the performance of diffusion models in a continual learning setup using the most common algorithms. We compare numerical results from running various CL methods in Section~\ref{sec:benchmark}. Next, we focus on specific phenomena, namely the poor quality of the bits-per-dimension metric in Section~\ref{sec:bpd} and timestep-dependent overfitting when using experience replay in Section~\ref{sec:overfitting}.

\textbf{Experimental setup} For our experiments, we use two datasets: MNIST and Fashion-MNIST. We divide the ten classes into five tasks, with two classes per task. We use a single DDPM model to learn all five tasks and pass the task ID during training and evaluation as a one-hot vector on which the model is conditioned. We train for 20K gradient descent steps in every task and adopt the hyperparameters of the diffusion model from \cite{ho2020denoising}.

\subsection{Comparing CL methods}
\label{sec:benchmark}
In this section, we provide an empirical comparison of selected continual learning methods, namely Finetuning, L2, Experience replay, and Generative replay. Table~\ref{tab:benchmark_mnist} and Table~\ref{tab:benchmark_fmnist} show the results for MNIST and Fashion-MNIST, respectively.

\textbf{Finetuning} maintains a high level of flexibility but quickly forgets how to generate samples from previous tasks, as indicated by the FID results. As discussed in Section~\ref{sec:bpd}, the decline is less evident regarding BPD. The \textbf{L2} method introduces strong regularization, which restricts forgetting and plasticity. Forgetting metrics for this method are generally close to zero and are the best among the compared methods, although the final average performance is not satisfactory, especially for the FID metric. 
In \textbf{Experience replay}, we apply a low coefficient for the replay loss compared to the main loss (coef. $= 0.01$). This successfully mitigates overfitting to the buffer data (see Section~\ref{sec:overfitting}), consequently leading to the best performance in terms of average FID and BPD. For completeness, we also present the result for equal coefficients (coef. $= 1$). 
We conducted a coefficient sweep for \textbf{Generative replay}, but the best performance was achieved for coef. $= 1$, underperforming Experience replay.

\begin{figure}
    \centering
    \vspace{-0.5cm}
    \begin{subfigure}[b]{0.49\textwidth}
        \centering
        \includegraphics[width=0.9\textwidth]{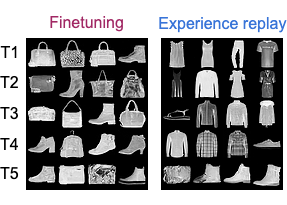}
        \caption{
    \label{fig:bpd_a} Samples generated from Finetuning and Experience replay (coef. $=1$). The former suffers from severe forgetting, at the end generating only classes from the last task; the latter successfully generates classes coming from respective tasks.}
    \end{subfigure}
    \hfill
    \begin{subfigure}[b]{0.49\textwidth}
        \centering
        \includegraphics[width=0.9\textwidth]{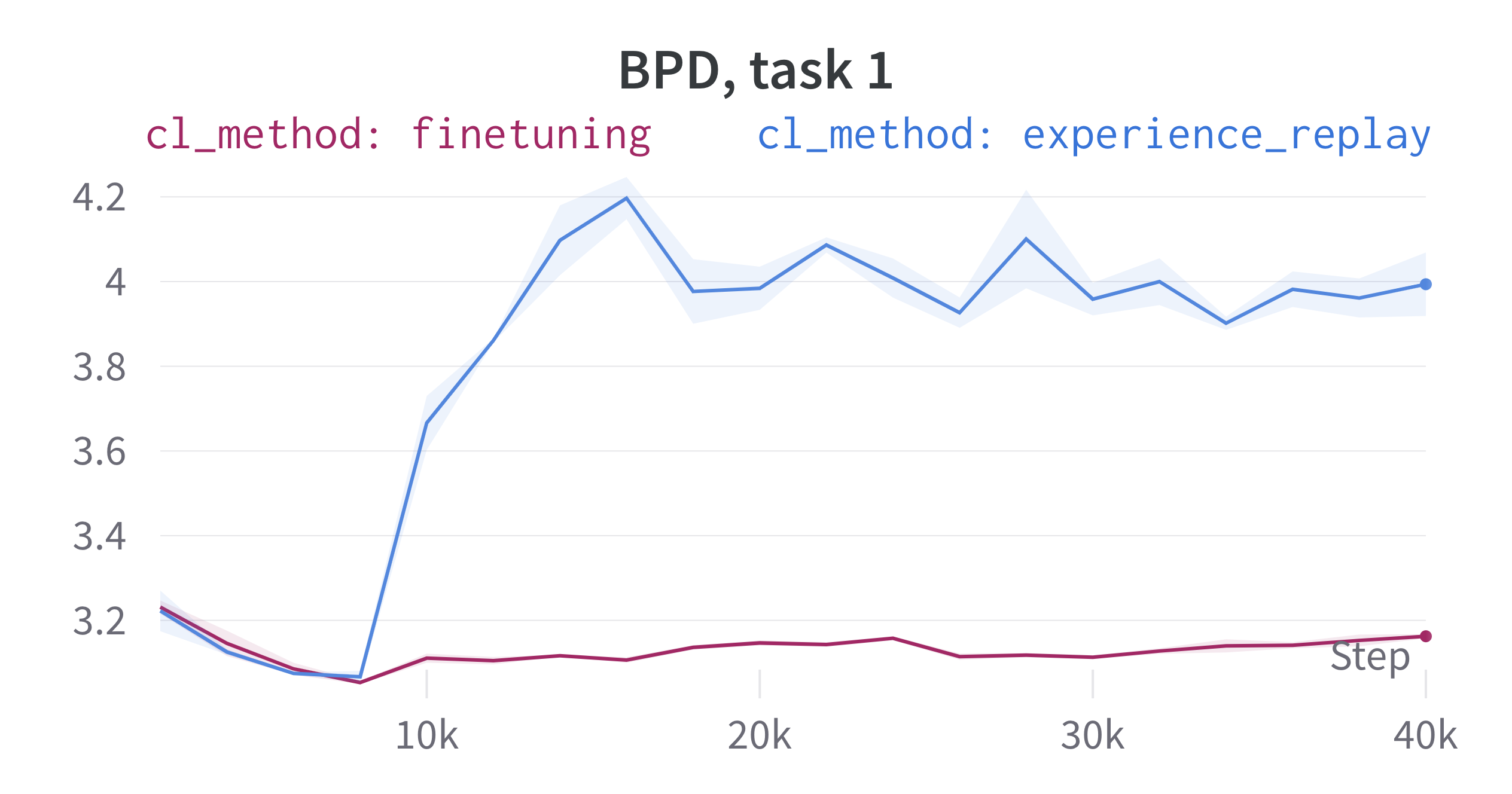}
        \caption{
    \label{fig:bpd_b} Even though Finetuning model forgets how to generate images from task 1, BPD on task 1 only slightly deteriorates, greatly outperforming Experience replay (coef. $=1$).}
    \end{subfigure}
    \hfill
    \begin{subfigure}[b]{0.49\textwidth}
        \centering
        \includegraphics[width=0.9\textwidth]{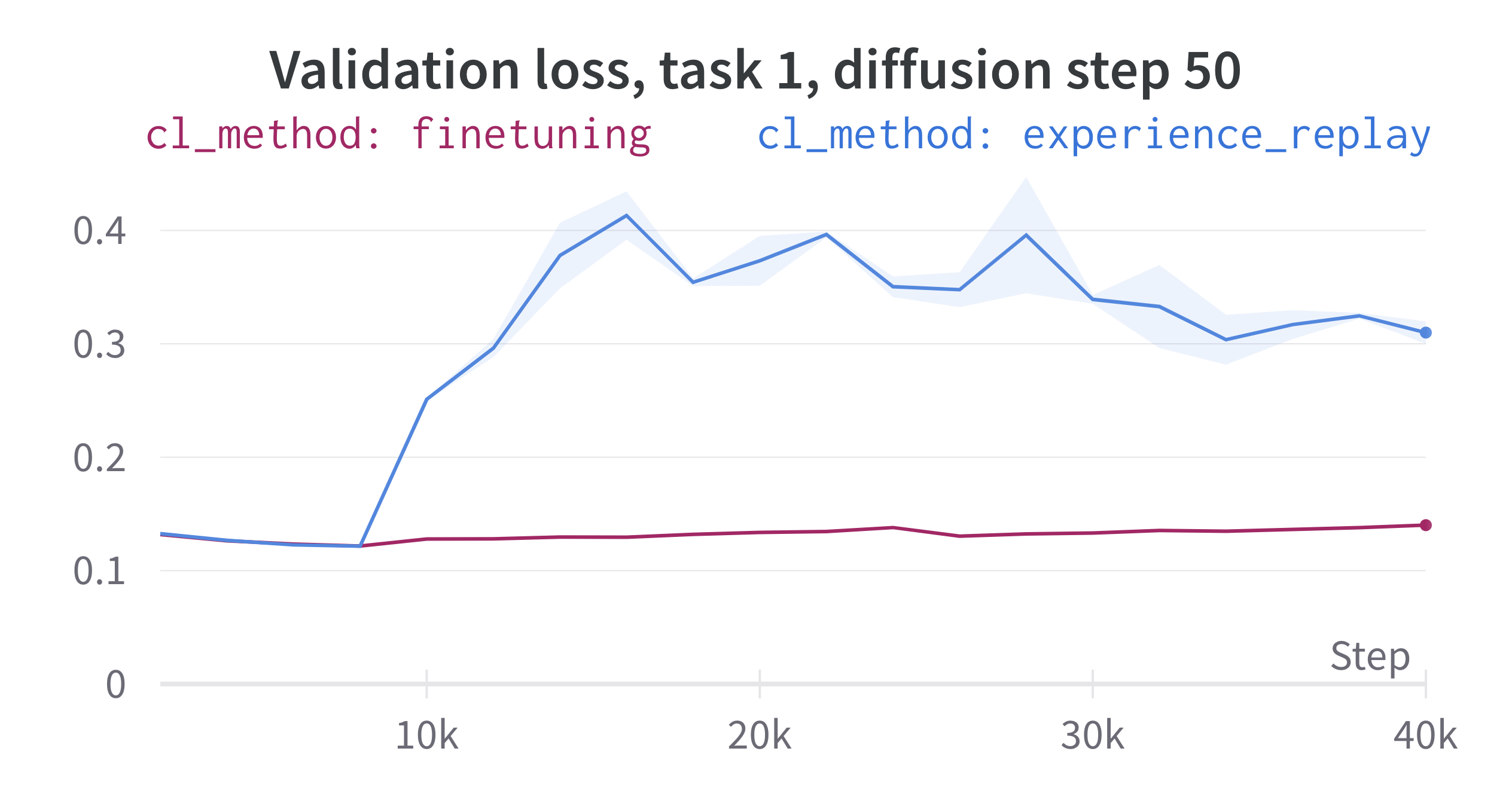}
        \caption{
    \label{fig:bpd_c} Part of the loss corresponding to early diffusion timestep is barely affected for Finetuning.}
    \end{subfigure}
    \hfill
    \begin{subfigure}[b]{0.49\textwidth}
        \centering
        \includegraphics[width=0.9\textwidth]{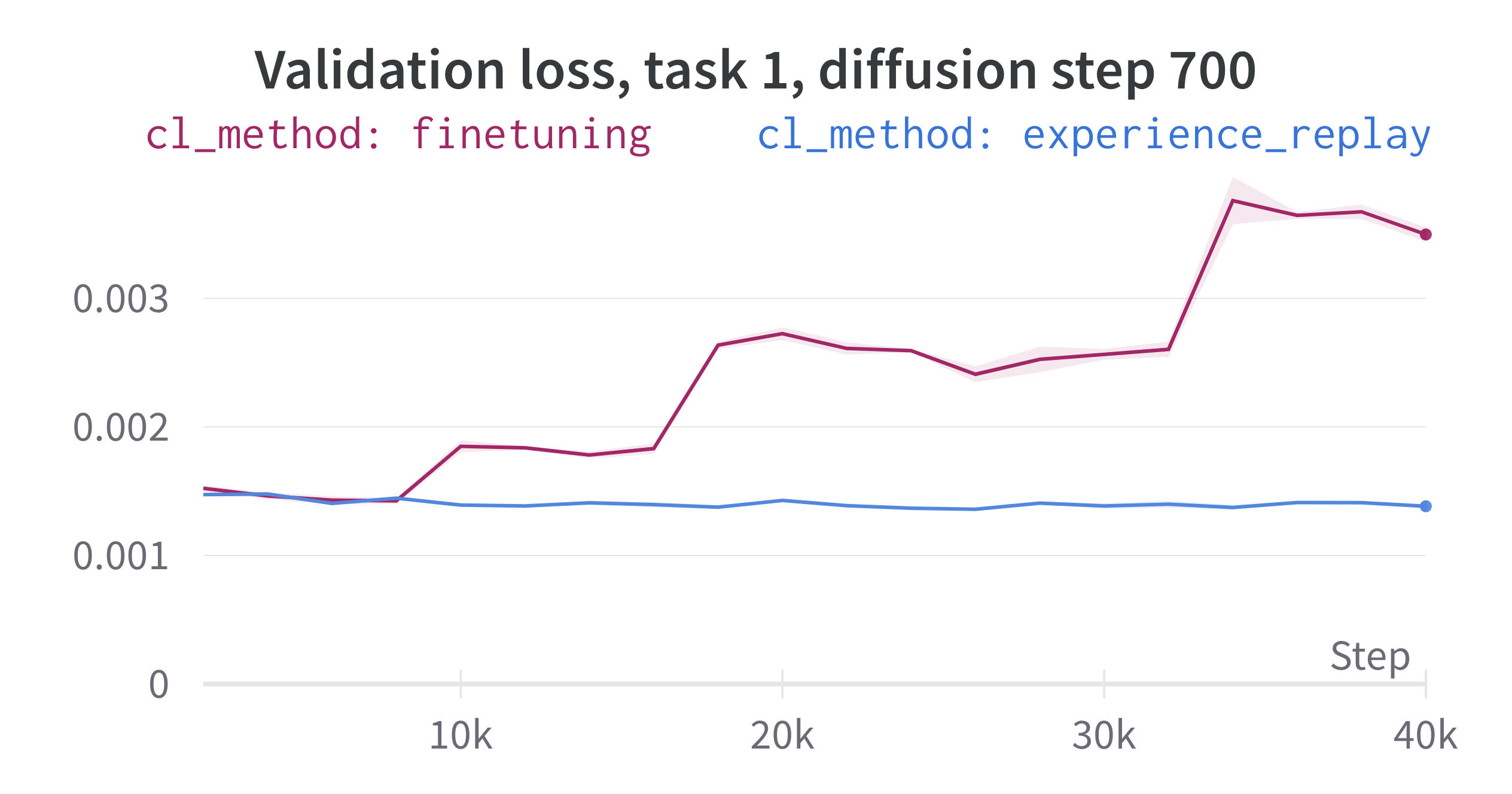}
        \caption{
    \label{fig:bpd_d} Part of the loss corresponding to late diffusion timestep deteriorates visibly for Finetuning. However, its scale is much smaller compared to (c) and thus it does not affect overall BPD much.}
    \end{subfigure}
    \vspace{-0.4cm}
\caption{Evolution of BPD values on task 1 for Finetuning and Experience replay methods during training on 5 subsequent tasks.}
\label{fig:bpd}
\end{figure}

\subsection{What metrics should be used? Pitfalls of BPD}
\label{sec:bpd}

Previous research has pointed out the limitations of log-likelihood as a metric and its inability to accurately reflect the capabilities of a generative model \cite{theis2015note, nalisnick2018deep}. In this section, we follow \cite{ho2020denoising,nichol2021improved} in calculating negative log-likelihood in bits-per-dimension (BPD). We demonstrate that this metric presents particular challenges in the context of generative continual learning and provide some insights into its unusual behavior in the case of diffusion models. 

To highlight the pitfalls of BPD, we calculate BPD values on examples from the first task while retraining the model with a simple finetuning method without any CL approach. Even though the model suffers from severe forgetting and does not generate images from previous tasks, only producing samples resembling the classes from the last task (as shown in Figure~\ref{fig:bpd_a}), the deterioration of the BPD metric is minor (see Figure~\ref{fig:bpd_b}). Surprisingly, the situation is reversed for the variant of Experience replay with rehearsal coef. $= 1$, where generations from the previous tasks remain plausible, but the loss deterioration is more visible.

To better understand the behavior of BPD, we propose examining its individual terms as it is a weighted sum of per-timestep quadratic terms (see Equations \ref{eq:sum} and \ref{eq:chunk}). We focus on a diffusion model with 1000 timesteps and observe that, at an early stage of the forward diffusion process (timestep 50), the Finetuning method shows minimal deterioration, as shown in Figure~\ref{fig:bpd_c}. However, at a later stage (timestep 700), deterioration is substantial (Figure~\ref{fig:bpd_d}). This indicates that the diffusion model forgets how to generate new image features from random noise while its denoising capabilities remain intact. The above observation is consistent with the work of \cite{deja2022analyzing}, which suggests that early diffusion timesteps correspond to a "denoising part" with good generalization to out-of-distribution data. Additionally, \cite[Figure 2]{nichol2021improved} shows that early diffusion timesteps contribute the most to log-likelihood. So the early timesteps, on which Finetuning's validation loss does not deteriorate significantly, are the ones that have the most impact on BPD, explaining the limited overall deterioration of BPD. However, it should be noted that the late timesteps, which contribute less to the BPD and are referred to as the "generative part" of the diffusion model in \cite{deja2022analyzing}, are crucial for generating high-quality samples. 

As a result, we caution against relying solely on variants of log-likelihood as a definitive measure for generative CL performance and advocate for reporting other metrics, such as FID.

\subsection{Experience replay: beware of overfitting}
\label{sec:overfitting}
Recent works~\cite{bonicelli2022effectiveness,verwimp2021rehearsal} have highlighted an important issue of models trained with replay methods overfitting to the examples stored in the rehearsal buffer. In this section, we demonstrate this phenomenon for diffusion models and further explore the specific structure of this overfitting.

\begin{figure}
\centering
    \begin{subfigure}[b]{0.49\textwidth}
        \centering
        \includegraphics[width=0.9\textwidth]{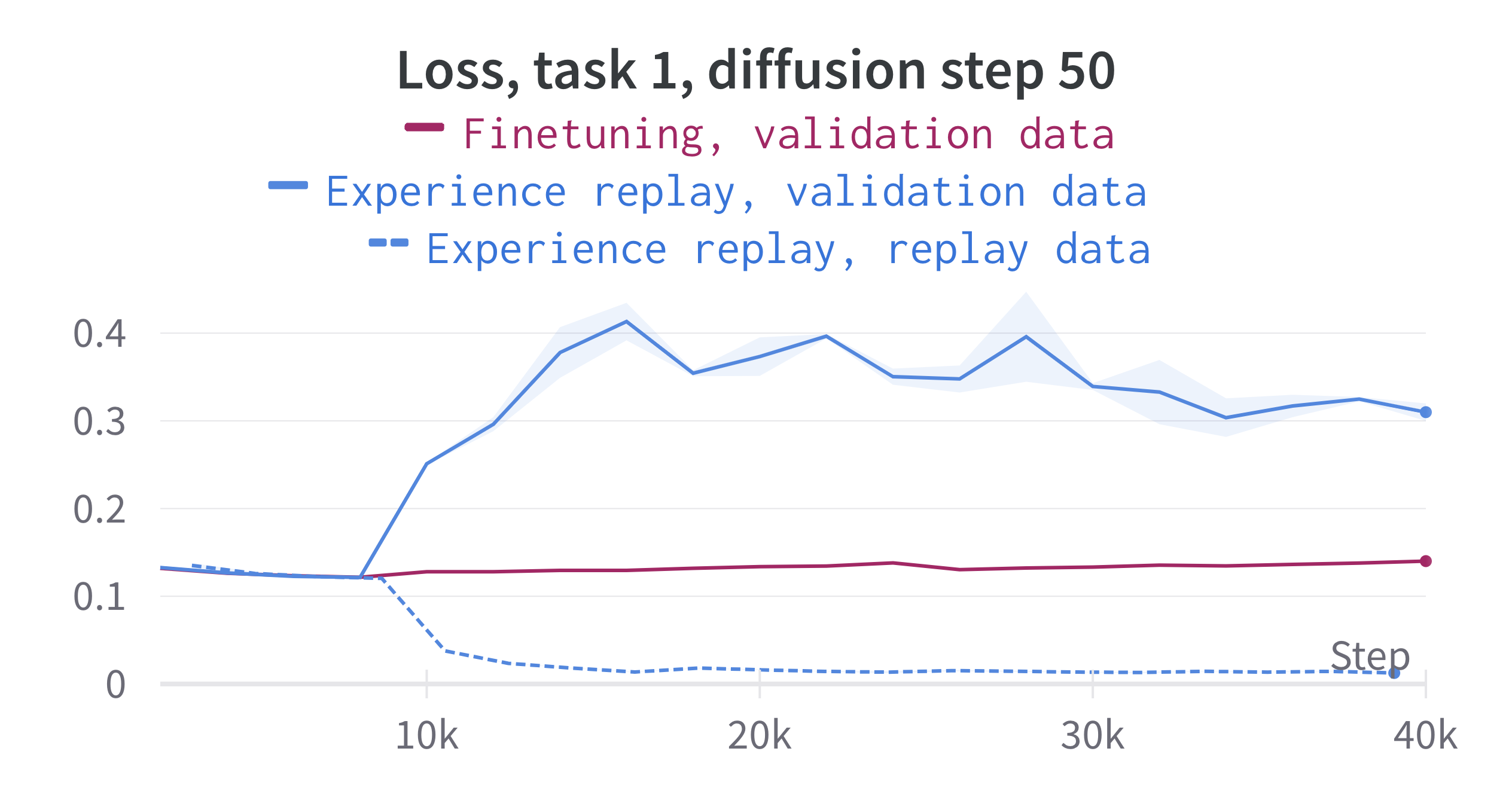}
        \caption{
\label{fig:overfitting_a} For the early diffusion timestep, overfitting on task 1 is clearly visible for the Experience replay (coef. $=1$) method.}
    \end{subfigure}
    \hfill
    \begin{subfigure}[b]{0.49\textwidth}
        \centering
        \includegraphics[width=0.9\textwidth]{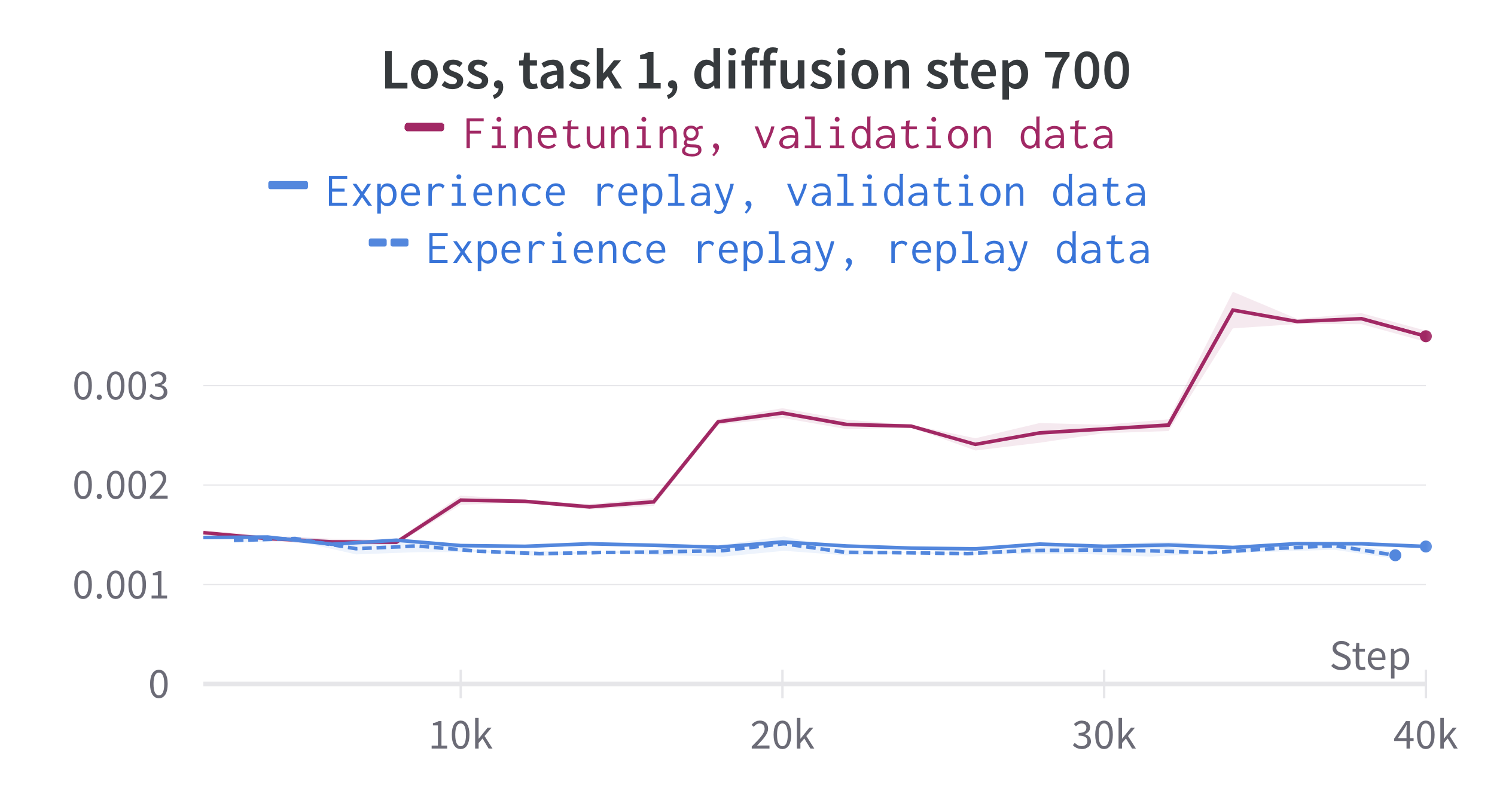}
        \caption{
\label{fig:overfitting_b} In contrast, there seems to be no overfitting for the late diffusion timestep.}
    \end{subfigure}
    \hfill
\caption{Demonstration of diffusion timestep-dependent overfitting for Experience replay (coef. $=1$).}
\label{fig:overfitting}
\end{figure}

In Figure~\ref{fig:bpd}, we saw that using replay can be detrimental to the validation loss of the replayed task. In our experiment, we employed a buffer size of 200 examples per task, which is relatively small, and assigned equal weights to the current task and rehearsal loss (rehearsal coef. $=1$). Therefore, overfitting was a likely cause for the loss deterioration. We demonstrate that this is indeed the case, and intriguingly, the extent of overfitting to the replay buffer heavily depends on the diffusion timestep. We present the results of the further investigation of the Experience replay (coef. $=1$) method in Figure~\ref{fig:overfitting}. For the early timestep ($50$), the loss of the first task on the replay data significantly drops after the task is finished, while the validation loss notably increases (Figure~\ref{fig:overfitting_a}). In contrast, for the late timestep ($700$), no overfitting occurs, resulting in Experience replay (coef. $=1$) outperforming the Finetuning baseline, as demonstrated in Figure~\ref{fig:overfitting_b}. We hypothesize that two primary factors contribute to this timestep-dependent overfitting: (1) the scale of the loss parts related to the early timesteps is greater, incentivizing the optimization process to overfit at these timesteps; and (2) in the late timesteps, more noise is introduced to the input before passing it to the neural network (see Equation~\ref{eq:loss_simplified}), acting as a natural regularizer that prevents overfitting. We believe that the described phenomenon is essential for
understanding the dynamics of continual learning of diffusion models. One possible direction of utilizing it would be to enhance the performance of the Experience replay method.

We note that reducing the relative weight for the rehearsal loss successfully addresses the overfitting problem, leading to significantly improved performance of the Experience replay (coef. $=0.01$) variant, see Section~\ref{sec:benchmark}.

\section{Conclusions and future work}

In this study, we have taken an initial step toward understanding the dynamics of continual learning in diffusion models. We provide benchmarking results for training diffusion models with various CL methods, demonstrating the strong performance of the experience replay with reduced rehearsal weight. We also present some qualitative observations regarding this framework, including the problematic behavior of the BPD metric and the occurrence of timestep-dependent overfitting when Experience replay (rehearsal coef. $=1$) is used.

We envision several interesting directions for future work. Firstly, we believe that novel CL strategies could be developed to enhance standard experience replay for diffusion models by leveraging their structural properties, such as the presence of diffusion timesteps. Secondly, extending the study to text-to-image and pretrained diffusion models would be also compelling. Lastly, providing more comprehensive benchmarking results, which include additional methods and datasets, would be beneficial to the research community.

\section*{Acknowledgements}

This research was supported by the PL-Grid Infrastructure.

{\small
\bibliographystyle{ieee_fullname}
\bibliography{biblio}
}

\end{document}